\documentclass[sigconf]{acmart}
\usepackage[utf8]{inputenc}
\usepackage{booktabs} % For formal tables
\usepackage{color}
\usepackage{hyperref}
\usepackage{url}
\usepackage{amsmath}
\usepackage{amsfonts}
\usepackage{graphicx}
\usepackage{bbm}
\usepackage{multirow, multicol}
\usepackage{comment}
\usepackage{xspace}
\usepackage{upquote}
\usepackage{float}
\usepackage{alltt}
\usepackage{algpseudocode}
\usepackage{pbox}
\usepackage{enumitem}
\usepackage{mathpartir}
\usepackage{wrapfig}
\usepackage{array}
\usepackage{hhline}
\usepackage{stmaryrd}
\usepackage{pifont}
\usepackage{makecell}
\usepackage{cuted}
\usepackage{subcaption}

\usepackage[ruled, vlined]{algorithm2e} % For algorithms

\SetAlFnt{\small}
\SetAlCapFnt{\small}
\SetAlCapNameFnt{\small}
\SetAlCapHSkip{0pt}
\SetArgSty{text}
\SetCommentSty{text}
\IncMargin{-\parindent}
\DontPrintSemicolon

\setcopyright{rightsretained}
\copyrightyear{2020}
\acmYear{2020}
\setcopyright{rightsretained}
\acmConference[CIKM '20]{Proceedings of the 29th ACM International Conference on Information and Knowledge Management}{October 19--23, 2020}{Virtual Event, Ireland}
\acmBooktitle{Proceedings of the 29th ACM International Conference on Information and Knowledge Management (CIKM '20), October 19--23, 2020, Virtual Event, Ireland}\acmDOI{10.1145/3340531.3411974}
\acmISBN{978-1-4503-6859-9/20/10}

\definecolor{comment-gray}{rgb}{0.5, 0.5, 0.5}

\SetCommentSty{mycommfont}

\definecolor{comment-red}{rgb}{0.8,0,0}
\definecolor{darkgreen}{rgb}{0,0.4,0}
\newcommand{\TODO}[1]{{\color{blue} TODO: #1}}
\newcommand{\Red}[1]{{\color{comment-red} #1}}
\newcommand{\Blue}[1]{{\color{blue} #1}}
\newcommand{\Green}[1]{{\color{darkgreen} #1}}

\newcommand{\SQA}[0]{Schema2QA\xspace}
\newcommand{\NLSchema}[0]{NLSchema\xspace}
\newcommand{\DBTalk}[0]{ThingTalk\xspace}
\newcommand{\cmark}{\ding{51}}%
\newcommand{\webqa}[0]{Schema2QA\xspace}
\newcommand{\webtalk}[0]{ThingTalk\xspace}

\newcommand{\QandA}[0]{Q\&A\xspace}
\usepackage{listings}
\usepackage{color}

\definecolor{dkgreen}{rgb}{0,0.6,0}
\definecolor{gray}{rgb}{0.5,0.5,0.5}
\definecolor{mauve}{rgb}{0.58,0,0.82}

\lstdefinelanguage{ThingTalk}{
  keywords={class, extends, skill},
  keywordstyle=\color{blue}\bfseries,
  ndkeywords={query, list},
  ndkeywordstyle=\color{darkgray}\bfseries,
  identifierstyle=\color{black},
  sensitive=false,
  comment=[l]{//},
  morecomment=[s]{/*}{*/},
  commentstyle=\color{purple}\ttfamily,
  stringstyle=\color{red}\ttfamily,
  morestring=[b]',
  morestring=[b]"
}

\begin{document}
\title[\webqa: High-Quality and Low-Cost Q\&A Agents for the Structured Web]{\webqa: High-Quality and Low-Cost Q\&A Agents \\ for the Structured Web}
%From Natural Language to Schema.org

\author{Silei Xu \quad Giovanni Campagna \quad Jian Li$^\dagger$ \quad Monica S. Lam}
\affiliation{%
  \institution{Computer Science Department}
  \institution{Stanford University}
  \institution{Stanford, CA, USA}
  \institution{\{silei, gcampagn, jianli19, lam\}@cs.stanford.edu}
}

\begin{abstract}
%!TEX root = ../paper.tex

% Virtual assistants have proprietary third-party skill platforms. 
% They train and own the voice interface to websites based on their submitted skill information. 
% This paper proposes \webqa, an open-source toolkit that leverages the Schema.org metadata found
% in many websites to automatically build skills.  \webqa has several advantages: 
% (1) \webqa is more accurate than commercial assistants in answering compositional queries involving multiple properties; 
% %(2) \webqa translates natural language into executable queries on the up-to-date data from the website;
% (2) it has a low-cost training data acquisition methodology that requires only writing a small number of annotations per domain and paraphrasing a small number of sentences.

Building a question-answering agent currently requires large annotated datasets, which are prohibitively expensive. This paper proposes \webqa, an open-source toolkit that can generate a Q\&A system from a database
schema augmented with a few annotations for each field. 
The key concept is to cover the space of possible compound queries on the database with a large number of in-domain questions synthesized with the help of a corpus of generic query templates.
The synthesized data and a small paraphrase set are used to train a novel neural network based on the BERT pretrained model. 

%The model can generalize a training set containing only synthesized and paraphrase data to understand real-world crowdsourced questions.

%By leveraging the popular Schema.org metadata format, many websites can quickly get their own Q\&A skill automatically using \SQA.    

We use \webqa to generate Q\&A systems for five Schema.org domains, restaurants, people, movies, books and music, and obtain an overall accuracy between 64\% and 75\% on crowdsourced questions for these domains.
Once annotations and paraphrases are obtained for a Schema.org schema, no additional manual effort is needed to create a Q\&A agent for any website that uses the same schema.  Furthermore, we demonstrate that learning can be transferred from the restaurant to the hotel domain, obtaining a 64\% accuracy on crowdsourced questions with no manual effort.
% and can answer questions that span three or more properties with 65\% accuracy.
\SQA achieves an accuracy of 60\% on popular restaurant questions that can be answered using Schema.org. Its performance is comparable to Google Assistant, 7\% lower than Siri, and 15\% higher than Alexa. 
It outperforms all these assistants by at least 18\% 
      on more complex, long-tail questions.

\iffalse
The open-source \webqa lets each website create and own its linguistic interface. 

It also lets search engines and aggregators query existing Schema.org data in websites.
Websites and search engines can make the interface available to any assistant, eliminating the lock-in of proprietary platforms. 
\fi

%technology keeps the linguistic web open by making these skills available to all assistants (and on the websites themselves) thus eliminating the lock-in by proprietary virtual assistant platforms.

%This paper shows we can make millions of websites available on voice assistants by using embedded Schema.org data in websites to answer site-specific questions directly.  

% \TODO{Can we do cross-domain -- such as ``with availability at noon today''.}

% \webqa is the first virtual assistant skill builder that uses a neural semantic parser to automatically translate complex natural language questions into executable queries.  We show we can teach the neural model compositionality across all the domains with relatively little real-user input by using domain-independent templates to generate a massive amount of synthetic training data.  Furthermore, accuracy can be incrementally improved easily by adding annotations to the definition of Schema.org to support natural language translation and writing templates to capture tricky domain-specific semantics.  

% is paper shows that skills built by \webqa can answer \TODO{many more complex queries than commercially available assistants}. 

\end{abstract}

%
% The code below should be generated by the tool at
% http://dl.acm.org/ccs.cfm
%

\iffalse


\fi

\fancyhead{}

\renewcommand{\authors}{Silei Xu, Giovanni Campagna, Jian Li, and Monica S. Lam}
\maketitle
\hypersetup{
  pdftitle={\webqa: High-Quality and Low-Cost Q\&A Agents for the Structured Web},
  pdfauthor={Silei Xu and Giovanni Campagna and Jian Li and Monica S. Lam},
  pdfkeywords={question answering; semantic parsing; KBQA; data synthesis; data augmentation; semantic web; linked data; Schema.org}
}

%!TEX root = ../paper.tex
\section{Introduction}
The adoption of virtual assistants is increasing at an unprecedented rate.
%About 3.25 billion virtual assistants are in use today globally and the number is estimated 
%to be 8 billion by 2023~\cite{voicebot2019-2}.
%Smart speakers alone have reached about 50 million American adults in just the first two years~\cite{techcrunch2018alexa}.  
Companies like Amazon and Google are rapidly growing their proprietary platforms of 
third-party skills, so consumers can access different websites and IoT devices by voice. 
Today, Alexa has 100,000 skills~\cite{voicebot2019} and Google claims 1M actions.
% The virtual assistant is a new entry point to the digital world. However, the world wide web is accessible by voice on just a few proprietary platforms.
 Websites can make themselves accessible
to each of the assistant platforms by supplying a skill containing many sample natural language invocations.
Virtual assistant platforms then use proprietary technology to train the linguistic interfaces. Such interfaces are unavailable outside the platforms, and reproducing them requires a prohibitively high investment, 
which is not affordable for all but the largest companies.
In the meantime, the existing proprietary linguistic interfaces can only accurately answer simple, popular questions. 
%Upon experimentation, we find that seemingly similar questions and more complicated questions cannot be answered correctly.

%Because of the high cost of data acquisition needed to develop natural language technology and the virality of network effect,
%we are observing that oligopolistic virtual assistants will control separate proprietary linguistic webs.,

We envision a future where cost-effective assistant technology is open and freely available so every company can create and own their voice interfaces.  Instead of submitting information to proprietary platforms, companies can publish it on their website in a standardized format, such as the Schema.org metadata.  The open availability of the information, together with the open natural language technology, enables the creation of alternative, competitive virtual assistants, thus keeping the voice web open.

Towards this goal, we have developed a new methodology and a toolkit, called \webqa, which makes it easy to create natural language \QandA systems. Developers supply only their domain information in database schemas, with a bit of annotation on each database field. \webqa creates an agent that uses a neural model trained to answer complex questions. \webqa is released as a part of the open-source Genie toolkit for virtual assistants\footnote{\url{https://github.com/stanford-oval/genie-toolkit}\\$^\dagger$Jian Li conducted this research at Stanford while visiting from the Computer Science and Engineering Department of the Chinese University of Hong Kong}.    

\subsection{A Hypothesis on Generic Queries}

Today's assistants rely on manually annotating real user data, which is prohibitively expensive for most companies. Wang et al. previously suggested a new strategy where canonical natural language questions and formal queries are automatically generated from a database schema~\cite{overnight}. These questions are then paraphrased to create a natural language training set. Unfortunately, this approach does not yield sufficiently accurate semantic parsers. We proposed substituting the single canonical form with developer-supplied templates to increase the variety of synthesized data, and training a neural network with both synthesized and paraphrased data~\cite{geniepldi19}. Good accuracy has been demonstrated on event-driven commands connecting up to two primitive functions.  However, such commands are significantly simpler than database queries. A direct application of this approach to \QandA is inadequate.

%E.g., ``why'' questions that require inferences are outside the scope of this paper. 
This research focuses on questions that can be answered with a database. 
Database systems separate {\em code} from {\em data}: with just a few operators in relational algebra, databases can deliver all kinds of answers by virtue of the data stored in the various fields of the tables.  Would there be such a separation of code from data in natural language queries that can be answered with a database? We define the part of the query that corresponds to the ``code'' as {\em generic}.  For example, ``what is the {\em field} of the {\em table}'' is a generic question that maps to the projection of a given {\em table} onto a given {\em field}.
%If so, can we just augment generic query knowledge with per-domain knowledge to answer questions in different domains effectively? 

We hypothesize that {\em by factoring questions into generic queries and domain-specific knowledge, we can 
greatly reduce the incremental effort needed to answer questions on new domains of knowledge.}  
We are {\em not} saying that every question can be derived by substituting the parameters in generic queries with the ontology of a domain. We only wish to reduce the need for annotated real data in training with a large corpus of synthesized sentences for each domain. 
The data synthesizer can generate hundreds of thousands of combinations of database operations and language variations to (1) provide better functional coverage, and (2) teach the neural network compositionality. 
To handle the full complexity and irregularity in natural language, we rely on the neural network to generalize from the combination of synthesized and human-produced sentences, using pretraining to learn general natural language.
% irrelevant???
% Note that this paper focuses on handling the text after it has been converted from speech; speech-to-text is outside the scope of this paper.  

\begin{figure}
\centering
\includegraphics[width=0.9\linewidth]{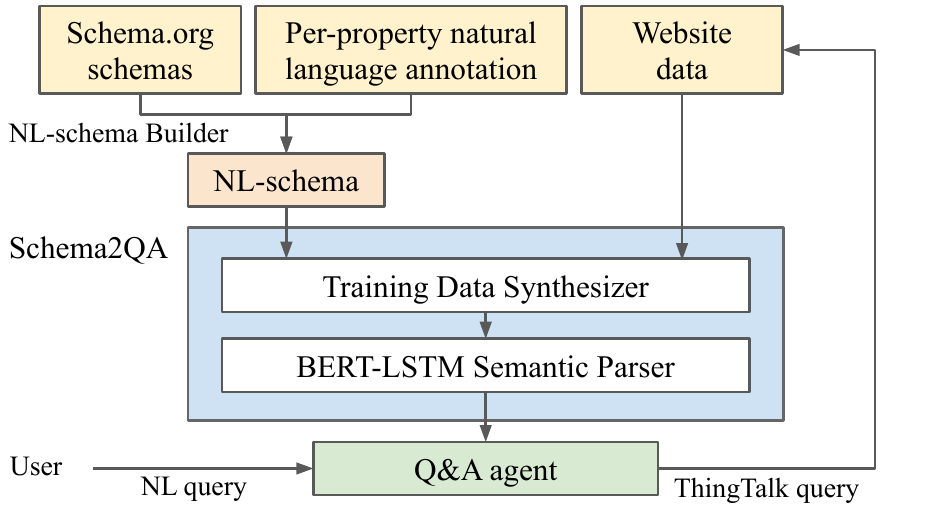}
\caption[]{The Schema2QA system architecture.}
\label{fig:overview}
\vspace{-0.5em}
\Description{}
\end{figure}

\subsection{Research Methodology}

Our research methodology is to design a representation that separates the generic and the domain-specific parts of a question. We extend the ThingTalk virtual assistant programming language~\cite{geniepldi19} to include the primitives necessary for querying databases.  We create Schema2QA, a tool that automatically generates a Q\&A agent from an {\em NL-schema}, a database scheme whose fields are annotated with ways they are referred to in natural language. 
%The agent can be entered in a repository such as Thingpedia~\cite{almondwww17} for use by a voice assistant.

\SQA builds the skill by training a novel neural semantic parser, based on the BERT pretrained model~\cite{Devlin2018Oct}, with mostly synthesized data.  It   
synthesizes pairs of natural-language sentences and ThingTalk code, from 
generic query templates and the domain-specific field annotations. 
For example, \SQA can synthesize ``Show me the address of 5-star restaurants with more than 100 reviews'' and this corresponding ThingTalk code:
\begin{align*}
&[\textit{address}]~\texttt{of}~\text{Restaurant}, \textit{aggregateRating.ratingValue} = 5 ~\texttt{\&\&}~\\
&\textit{aggregateRating.reviewCount} \ge 100
\end{align*} 
It augments the synthesized data and a small number of paraphrased sentences with real parameter values. 
At inference time, the parser translates questions into ThingTalk, which is executed to retrieve the answer from a database, which may be updated dynamically. 

One important ontology is Schema.org, a commonly used representation for structured data in web pages.  We create an NL-schema Builder tool that converts the RDF-based Schema.org representation into an NL-schema, which is openly available in Thingpedia~\cite{almondwww17}. Once manual annotations and paraphrases are available for a domain, a Q\&A agent can be generated automatically for any website in that domain using its Schema.org metadata. The agent can also be entered in Thingpedia, an open skill repository used currently by the open-source privacy-preserving Almond assistant~\cite{almondwww17}. The architecture of \SQA is shown in Fig.~\ref{fig:overview}.

% dangerous claim? we don't evaluate it 
% We can also apply \webqa to all the domains in Schema.org to answer all the questions possible on the structured web.

%This enables \webqa to answer questions that popular commercial assistants cannot, as shown by the examples in Table~\ref{table:comparison}.

\subsection{Contributions}
The contributions of this paper include:
\begin{itemize}
\item A new methodology and a \webqa toolkit, that significantly reduce the cost in creating \QandA systems for databases by leveraging a set of generic question templates. 
      The toolkit synthesizes a large training set cost-effectively using a template-based algorithm,
      and trains 
      a novel BERT-based neural semantic parser. %The parser can generalize to questions and entities not seen during training.

%\item A high-level query language, called \webtalk, which simplifies the representation of complex 
%      questions and is designed to improve the accuracy of natural language translation.
\item 
We demonstrated that \webqa is applicable to Schema.org on six domains: restaurants, people, books, movies, music, and hotels. We constructed a large training set
      of complex web queries, annotated in ThingTalk, with more than 400,000 questions in each domain. 
      This dataset also includes 5,516 real-world crowdsourced questions. This test set can serve as a benchmark for future \QandA systems\footnote{The data can be downloaded from \url{https://stanford-oval.github.io/schema2qa}}.

\item Experimental results show that \QandA systems built with \webqa can
      understand a diverse set of complex questions, with an average query accuracy of 70\%.
      We also show that we can transfer the restaurant skill to the hotel domain with no manual effort, achieving 64\% accuracy on crowdsourced questions.  
      %In comparison, the top 3 commercial virtual assistants achieve an answer accuracy (an upper bound for query accuracy) between 36\% and 53\% on restaurants. 

\item 
  \SQA achieves an accuracy of 60\% on popular restaurant questions that can be answered using Schema.org. The accuracy is comparable to Google Assistant, 7\% lower than Siri and 15\% higher than Alexa. 
  In addition, \SQA outperforms all these assistants by at least 18\% 
      on more complex, long-tail questions.
\end{itemize}

\subsection{Outline}

We first discuss related work in Section~\ref{sec:related-work}.
%Section~\ref{sec:overview} presents an overview of the system. 
We present how we synthesize varied questions along with their ThingTalk representation, the application to Schema.org, and our neural model in Sections~\ref{sec:nl-to-dbtalk} through~\ref{sec:model}, respectively. 
Lastly, we present experimental results and conclude. 

%!TEX root = ../paper.tex
\section{Related Work}
\label{sec:related-work}

\paragraph{Question answering}
Question answering (QA) is a well-studied problem in
natural language processing. %, with work dating back to the 60s~\cite{green1961baseball}.
A subset of the QA field is to build Natural language Interfaces to Databases (NLIDB).
Early proposed systems to solve this problem 
use rule-based approaches,
which are not robust to variations in natural language
~\cite{popescu2003towards, giordani2008mapping, li2014constructing}.
% A comprehensive review of the question answering literature can be found in ~\cite{}.
More recently, 
neural semantic parsing has been widely adopted for question answering~\cite{
%zelle1994inducing, zelle1996learning, tang2001using, zettlemoyer2005learning, wong2007learning, yahya2012natural, pasupat2015compositional, overnight,
dong2016language, jia2016data}.
%which generate executable queries from natural language questions directly.
However, this technique requires a large corpus of annotated questions, which is expensive. 
Previous work has proposed crowdsourcing paraphrases to bootstrap new semantic
parsers~\cite{overnight}. Our work on Genie further suggested training with data
synthesized from manually tuned templates, based on the constructs in a programming language
where each skill provides domain-specific templates mapping to website-specific APIs~\cite{geniepldi19}.
DBPal~\cite{weir2020dbpal} uses a similar approach to augment existing datasets with synthesized data. 
In this paper, we propose a more varied set of templates, covering not only the variety in functionality, but also
the variety in natural language. This avoids the need for domain-specific templates and existing datasets.
%Furthermore, by leveraging Schema.org markup, the effort in building skills with
%\webqa is per-domain rather than per-skill.
%by leveraging existing structured and unstructured web information.
% that is synthesized We extend on that line of work by showing 

\iffalse
For this reason, the largest human-written semantic parsing dataset is the WikiSQL dataset~\cite{zhong2017seq2sql},
with about 80,000 questions. WikiSQL was obtained using the paraphrasing technique~\cite{overnight}, in which sentence and query pairs are
sampled from hand-written rules and paraphrased by crowdsource workers. Previous work extended on this technique and proposed the Genie
toolkit~\cite{geniepldi19}, which introduces a template language so engineers can refine the synthesized question-query pairs, and
combines paraphrased and synthesized data during training. In this paper, we extend Genie to reduce the hand engineered effort,
by leveraging existing structured and unstructured web information.
\fi

\paragraph{Semantic parsing}
Recent work in semantic parsing has focused on generating SQL directly from natural language queries~\cite{zhong2017seq2sql, xu2017sqlnet, yavuz-etal-2018-takes, DBLP:journals/corr/abs-1809-08887}.
Most of the work employs neural encoder-decoder frameworks~\cite{dong2016language, iyer2017learning, McCann2018decaNLP} with attention.
Later work on the WikiSQL dataset~\cite{zhong2017seq2sql} leveraged the constrained space of questions in that dataset and developed syntax-specific
decoding strategies~\cite{yu2018typesql,yin2018tranx,wang2018robust}. 
Often, these models use pre-trained models as embeddings to a larger, SQL-aware encoder~\cite{guo2019towards}. \citeauthor{hwang2019comprehensive}~\cite{hwang2019comprehensive}
explored different combinations of the BERT encoder with LSTM decoders: a vocabulary-only decoder and a pointer-only one.
After finding both to be ineffective, they proposed an application-specific decoder. Our model instead uses a pointer-generator controlled
by a switch probability. This is effective in our task, and it is also simpler (thus easier to train) and more general.

\paragraph{Using Schema.org in virtual assistants}
The Schema.org vocabulary is in active use by commercial virtual assistants for their builtin skills, for example
in the Alexa Meaning Representation Language~\cite{kollar2018alexa} and in Google Assistant. Compositional queries based on Schema.org require expert annotation
on large training sets~\cite{perera2018multi}. Furthermore, because of the annotation cost, compositional query capabilities are not available to third-parties,
which are limited to an intent classifier~\cite{dblp:journals/corr/abs-1711-00549}.
%Google Assistant is also able to automatically generate skills for websites that use schema.org markup, and supports five domains.
\iffalse
Each skill is automatically built by pairing the crawled website data with predefined models. While this approach supports multiple websites,
it requires a substantial amount of work in annotating the training set, and the models are not transferrable. In addition, automatically generated skills
do not answer aggregated questions.
\fi
Our approach only requires a small amount of developer effort, and the effort can be shared among all websites using the same Schema.org properties. Furthermore, each website
can own their generated semantic parser and improve it for their own use case, instead of relying on a proprietary one.

%\input{sections/overview}
%!TEX root = ../paper.tex
\section{Natural Language to \DBTalk}
\label{sec:nl-to-dbtalk}

The principle behind neural semantic parsers is to let the neural network generalize from many possible sentences annotated with their meaning in a formal representation. However, because languages are compositional, the neural network must see many possible combinations in order not to overfit. In contrast, compositionality is baked into grammar-based parsers, but they fail to handle the irregularities and contextual variations in natural language.  {\em We propose to use templates to generate a large variety of perfectly annotated natural language sentences to provide coverage and to teach the neural network compositionality, and we use a small number of paraphrases to expose the network to more variety in natural language.} Our approach allows the network to generalize and understand natural language with significantly less manually labeled data. 

\setlength\tabcolsep{3pt}
\begin{table*}
\caption{Canonical templates mapping natural language to \DBTalk code.}
\vspace{-0.5em}
\footnotesize
\centering
\begin{tabular}{p{1.4cm}|p{0.84cm}p{5.4cm}|p{3.7cm}|p{5.3cm}}
\toprule
{\bf Operator}               & \multicolumn{2}{l|}{\bf Natural language template}           & {\bf \DBTalk}                                       & \bf Minimal Example \\
\hline
\multirow{4}{*}{Selection}   & table $:=$& $t : \textit{table}$ with $f : \textit{fname}$ equal to $v : \textit{value}$     & $t, f = v$   & [restaurants] with [cuisine] equal to [Chinese] \\ 
                             & $\quad\quad\;\,\vert$& $t : \textit{table}$ with $f : \textit{fname}$ greater than $v : \textit{value}$ & $t, f \ge v$ & [restaurants] with [rating] greater than [3.5]  \\
                             & $\quad\quad\;\,\vert$& $t : \textit{table}$ with $f : \textit{fname}$ less than $v : \textit{value}$    & $t, f \le v$ & [restaurants] with [rating] less than [3.5]  \\
                             & $\quad\quad\;\,\vert$& $t : \textit{table}$ with $f : \textit{fname}$ containing $v : \textit{value}$   & $t, \text{contains}(f, v)$ & [people] with [employers] containing [Google]  \\
\hline
Projection                   & fref $:=$& the $f : \textit{fname}$ of $t : \textit{table}$                    & $f~\texttt{of}~t$         & the [cuisine] of [restaurants] \\
\hline
\multirow{1}{*}{Join}        & table $:=$& the $t_1 : \textit{table}$ of $t_1 : \textit{table}$           & $(t_2~\texttt{join}~t_1), \text{in\_array}(id, t_2)$ & [reviews with ...] of [restaurants with ...]     \\
\hline
\multirow{2}{*}{Aggregation} & fref $:=$& the number of \textit{table}                     & $\texttt{aggregate}~\text{count}~\texttt{of}~t$ & the number of [restaurants] \\
                             & $\quad\quad\;\,\vert$& the \textit{op} $f : \textit{fname}$ in $t : \textit{table}$               & $\texttt{aggregate}~\textit{op}~f~\texttt{of}~t$ & the [average] [rating] of [restaurants] \\
\hline
\multirow{4}{*}{Ranking}     & table $:=$& the $t : \textit{table}$ with the min $f : \textit{fname}$      & $(\texttt{sort}~f~\texttt{asc}~\texttt{of}~t)[1]$  & the [restaurants] with the min [rating] \\
                             & $\quad\quad\;\,\vert$& the $t : \textit{table}$ with the max $f : \textit{fname}$      & $(\texttt{sort}~f~\texttt{desc}~\texttt{of}~t)[1]$ & the [restaurants] with the max [rating] \\
                             & $\quad\quad\;\,\vert$& the $n : \textit{number}$ $t : \textit{table}$ with the min $f : \textit{fname}$  & $(\texttt{sort}~f~\texttt{asc}~\texttt{of}~t)[1:n]$ & the [3] [restaurants] with the min [rating] \\
                             & $\quad\quad\;\,\vert$& the $n : \textit{number}$ $t : \textit{table}$ with the max $f : \textit{fname}$  & $(\texttt{sort}~f~\texttt{desc}~\texttt{of}~t)[1:n]$ & the [3] [restaurants] with the max [rating] \\
\hline
\multirow{2}{*}{Quantifiers} & table $:=$& $t_1 : \textit{table}$ with $t_2 : \textit{table}$             & $t_1, \texttt{exists}(t_2, \text{in\_array}(id, t_1))$ & [restaurants with ...] with [reviews with ...] \\
                             & $\quad\quad\;\,\vert$& $t_1 : \textit{table}$ with no $t_2 : \textit{table}$          & $t_1, !\texttt{exists}(t_2, \text{in\_array}(id, t_1))$ & [restaurants with ...] with [no reviews with ...] \\
\hline
\multirow{2}{1.5cm}{Row-wise function} & fref $:=$& the distance of $t : \textit{table}$ from $l : \textit{loc}$        & $\texttt{compute}~\text{distance}(\textit{geo}, l)~\texttt{of}~t$ & the distance of [restaurants] from [here] \\
                             & $\quad\quad\;\,\vert$& the number of $f : \textit{fname}$ in $t : \textit{table}$          & $\texttt{compute}~\text{count}(f)~\texttt{of}~t$ & the number of [reviews] in [restaurants] \\
\bottomrule
\end{tabular}
\label{table:db-ops}
\end{table*}

Here we first define \DBTalk, the formal target language of our semantic parser.  We then describe the natural language sentences we synthesize with (1) canonical templates to cover all the possible queries and (2) generic query templates to provide linguistic variety. 

%!TEX root = ../paper.tex
\begin{figure}
\footnotesize
\vspace{-1em}
\begin{tabbing}
123\=123456789012345678901\=\kill\\
\>Table $t$\>$\textit{tn}~~\vert~~\textit{sel}~~\vert~~\textit{pr}~~\vert~~\textit{agg}~~\vert~~\textit{cmp}~~\vert~~\textit{sort}~~\vert~~\textit{idx}~~\vert~~\textit{join}$\\
\>Selection $\textit{sel}$\>$t, f$\\
\>Projection $\textit{pr}$\>$\texttt{[}\textit{fn}^+\texttt{]}~~\texttt{of}~~t$\\
\>Aggregation $\textit{agg}$\>$\texttt{aggregate}~~\textit{aggop}~~\texttt{of}~~t$\\
\>Computation $\textit{cmp}$\>$\texttt{compute}~~\textit{expr}~~\left\{\texttt{as}~~\textit{fn}\right\}^?~~\texttt{of}~~t$\\
\>Sorting $\textit{sort}$\>$\texttt{sort}~~\textit{fn}~~\left\{\texttt{asc}~~\vert~~\texttt{desc}\right\}~~\texttt{of}~~t$\\
\>Indexing $\textit{idx}$\>$t~\texttt{[}v\texttt{]}~~\vert~~t~\texttt{[}v : v\texttt{]}$\\
\>Join $\textit{join}$\>$t~~\texttt{join}~~t$\\
\>Filter $f$ \>$\texttt{true}~~\vert~~\texttt{false}~~\vert~~\texttt{!}\textit{f}~~\vert~~\textit{f}\texttt{ \&\& }\textit{f}~~\vert~~
\textit{f}~\texttt{||}~\textit{f}~~\vert$\\
        \>\>$v~\textit{cmpop}~v~~\vert~~t~~\texttt{\{}~~f~~\texttt{\}}$\\
\>Expression $\textit{expr}$ \> $v~~\vert~~\textit{expr}~\texttt{+}~\textit{expr}~~\vert~~\textit{expr}~\texttt{-}~\textit{expr}~~\vert~~\textit{expr}~\texttt{*}~\textit{expr}~~\vert$\\
        \>\>$\textit{expr}~\texttt{/}~\textit{expr}~~\vert~~\texttt{distance}\left(\textit{expr}, \textit{expr}\right)~~\vert~~\textit{aggop}\left(v\right)$\\
\>Comparison $\textit{cmpop}$ \> $=~~\vert~~\ge~~\vert~~\le~~\vert~~=\sim\vert~~
\texttt{contains}~~\vert~~\texttt{in\_array}~~\vert~~\ldots$\\
\>Agg. operator $\textit{aggop}$ \> $\text{count}~~\vert~~\text{sum}~~\vert~~\text{avg}~~\vert~~\text{min}~~\vert~~\text{max}$\\
\>Table name $\textit{tn}$\>identifier\\
\>Field name $\textit{fn}$\>identifier\\
\>Value $v$ \>$\text{literal}~~\vert~~\textit{fn}~~\vert~~\texttt{lookup}(\text{literal}, \textit{tn})$
\end{tabbing}
\vspace{-1em}
\caption{The formal definition of \DBTalk.}
\vspace{-1em}
\label{fig:DBTalk-grammar}
\end{figure}

\iffalse
Previous work 
has shown it's challenging to translate natural language into SQL directly
\cite{zhong2017seq2sql, xu2017sqlnet, iyer2017learning, DBLP:journals/corr/abs-1809-08887, yavuz-etal-2018-takes},
and it's important that the formal representation resembles natural language~\cite{geniepldi19}.
We design a formal query language called \DBTalk
to represent complex queries and at the same time facilitate both the synthesis of the training set, and the translation 
from natural language with a neural model.
\fi

\subsection{The \DBTalk Query Language}
\label{sec:query-language}

Our target language is an extension of ThingTalk, a previously proposed programming language optimized for translation
from natural language~\cite{geniepldi19}. Our extensions make ThingTalk a functional subset of SQL. In this paper,
we focus on the query aspects of ThingTalk; readers are referred to previous work for the design of the rest of the language. 
\DBTalk is designed with a relational database model.   
\DBTalk queries have the form:
\begin{align*}
\left[\textit{fn}^+~~\texttt{of}\right]^?~~\textit{table}~\left[, \textit{filter}\right]^? \left[\texttt{join}~~\textit{table}~\left[, \textit{filter}\right]^?\right]^*
\end{align*}
where \textit{table} is the type of entity being retrieved (similar to a table name
in SQL), \textit{filter} applies a selection predicate that can make use of the fields
in the table, and \textit{fn} is an optional list of field names to project on.
The full grammar is shown in Fig.~\ref{fig:DBTalk-grammar}.
\DBTalk queries support the standard relational algebra operators commonly
used by natural language questions. These include
sorting a table, indexing \& slicing a table, aggregating all results,
and computing a new field for each row.
The \texttt{join} operator produces a table that contains the fields from both tables; fields from the second table shadow the fields of the first table.
All the operators can be combined compositionally. 

\DBTalk uses a static type system. It has native support for named entities, such as people, brands, countries, etc.
Every table includes a unique \textit{id} field, and \DBTalk uses the ``lookup'' operator to look up the ID of a specific entity by name.
\DBTalk introduces array types to express joins at a higher level, and to avoid the use of glue tables
for many-to-many joins.
\DBTalk also includes common types, such as locations, dates and times,
and also includes user-relative concepts such as ``here'' and ``now''. The latter allows the parser to translate a natural language sentence containing those words to a representation
that does not change with the current location or current time.

For example, the \texttt{distance} operator can be
used to compute the distance between two locations. Combined
with sorting and indexing, we can express the query ``find the nearest
restaurant'' as:
\begin{tabbing}
(\texttt{sort}~\textit{distance}~\texttt{asc}~\texttt{of}~\texttt{comp}~\text{distance}(\textit{geo}, \texttt{here})~\texttt{of}~\text{Restaurant})[1]
\end{tabbing}
The query reads as: select all restaurants, compute the distance between the
field \textit{geo} and the user's current location (and by default, store it
in the \textit{distance} field), sort by increasing distance, and then choose
the first result (with index 1).

\DBTalk is designed such that the clauses in the query compose in the same way as the phrases
in English. This helps with synthesis. For example, the query ``who wrote the 1-star review for Shake Shack?''
is expressed as:
\begin{tabbing}
123456789012\=12345\=\kill
$[~\textit{author}~]~\texttt{of} ~~~((\text{Restaurant}, \textit{id} = \texttt{lookup}(\text{``shake shack''}))$\\
\>$\texttt{join} (\text{Review}, \textit{reviewRating.ratingValue} = 1)),$\\
\>$\texttt{in\_array}(\textit{id}, \textit{review})$
\end{tabbing}
The query reads as ``search the restaurant `Shake Shack', do a cross-product with all 1-star reviews, and then
select the reviews that are in the list of reviews of the restaurant; of those, return the author''.
% In this query, the first mention of \textit{id} is the ID of the
% ``Restaurant'' table, and the second one is the ID of the ``Review'' table. 
%
%  Hence,
% in the ``in\_array'' operation, \textit{id} is a field of ``Review'', and \textit{review} is a field of ``Restaurant''.
 The ``1-star review'' phrase corresponds to the ``Review'' clause of the query,
and ``Shake Shack'' corresponds to the ``Restaurant'' clause. 
The combination ``the 1-star review for Shake Shack'' corresponds to 
the join and ``in\_array'' selection expression. Adding ``who wrote'' to the sentence is equivalent to
projecting on the ``author'' field.

\subsection{Canonical Templates}

\SQA uses the previously proposed concept of \textit{templates}
to associate \DBTalk constructs with natural language~\cite{geniepldi19}.
Templates are production rules mapping the grammar of natural language to a \textit{semantic function} that produces the corresponding code. 
Formally, a template is expressed as:
\begin{align*}
\textit{nt} := \left[v : \textit{nt}~~\vert~~\text{literal}\right]^+ \Rightarrow \textit{sf}
\end{align*}
The non-terminal \textit{nt} is produced by expanding the terminals and non-terminals
on the right-hand side of the $:=$ sign.
The bound variables $v$ are used by the semantic function \textit{sf} to produce the corresponding code.

For each operator in \DBTalk, we can construct a \textit{canonical template} expressing that operator in natural language.
The main canonical templates are shown in Table~\ref{table:db-ops}. The table omits logical connectives such as ``and'', ''or'', and ``not'' for brevity.
These canonical templates express how the composition of natural language is reflected in the composition of corresponding relational algebra operators to form complex queries.

With the canonical templates, we can cover the full functionality of \DBTalk, and thus
the full span of questions that the system can answer. However, these synthesized sentences do not reflect how people actually ask questions. In the next section, we discuss how we can increase the
variety in the synthesized set with a richer set of generic templates.

\begin{table*}
\vspace{-0.5em}
\caption{Variety in question structure given the fact that Dr. Smith is Ann's doctor}%  , assuming a table of patients with a ``doctor'' field}.
\vspace{-0.5em}
\fontsize{7.7}{10}\selectfont
\begin{tabular}{lllll}
\toprule
{\bf \Red{Relation}} & {\bf Part-of-Speech} & {\bf Statement} & \bf Unknown: \Green{Ann} & \bf Unknown: \Blue{Dr. Smith}\\
\midrule
\Red{Doctor} & has-a noun & \Green{Ann} has \Blue{Dr. Smith} \Red{as a doctor}. & \Green{Who} has \Blue{Dr. Smith} \Red{as a doctor}?                    & \Blue{Who} does \Green{Ann} have \Red{as a doctor}? \\
&is-a noun  & \Blue{Dr. Smith} is \Red{a doctor of} \Green{Ann}.  & \Green{Who} is \Blue{Dr. Smith} \Red{a doctor of}?                      & \Blue{Who} is \Red{a doctor of} \Green{Ann}? \\
&active verb       & \Blue{Dr. Smith} \Red{treats} \Green{Ann}.             & \Green{Whom} does \Blue{Dr. Smith} \Red{treat}?                            & \Blue{Who} \Red{treats} \Green{Ann}? \\
&passive verb & \Green{Ann} is \Red{treated by} \Blue{Dr. Smith}.      & \Green{Who} is \Red{treated by} \Blue{Dr. Smith}?                       & \Red {By} \Blue{whom} is \Green{Ann} \Red{treated}? \\
\midrule
\Red{Patient}&has-a noun & \Blue{Dr. Smith} has \Green{Ann} \Red{as a patient}.  & \Green{Who} does \Blue{Dr. Smith} have \Red{as a patient}?                & \Blue{Who} has \Green{Ann} \Red{as a patient}? \\
&is-a noun  & \Green{Ann} is \Red{a patient of} \Blue{Dr. Smith}.   & \Green{Who} is \Red{a patient of} \Blue{Dr. Smith}?                    & \Blue{Who} is \Green{Ann} \Red{a patient of}? \\
&active verb       & \Green{Ann} \Red{consults with} \Blue{Dr. Smith}.           & \Green{Who} \Red{consults with} \Blue{Dr. Smith}?                            & \Red{With} \Blue{whom} does \Green{Ann} \Red{consult}? \\
&passive verb & \Blue{Dr. Smith} is \Red{consulted by} \Green{Ann}.        & \Red{By} \Green{whom} is \Blue{Dr. Smith} \Red{consulted}?                            & \Blue{Who} is \Red{consulted by} \Green{Ann}? \\
\bottomrule
\end{tabular}
\label{table:question_types2}
\end{table*}
%!TEX root = ../paper.tex
\begin{table*}
\centering
\caption{Example annotations for fields servesCuisine, rating, and alumniOf in different parts of speech.}
\vspace{-0.5em}
\small
\begin{tabular}{lllll}
\toprule
{\bf POS}    & {\bf servesCuisine : String}                    & {\bf rating : Number}                             & {\bf alumniOf : Organization} & {\bf Example sentence for alumniOf}\\
\hline
has-a noun   & $\textit{value}$ cuisine, $\textit{value}$ food & rating $\textit{value}$, $\textit{value}$ star    & $\textit{value}$ degree, alma mater $\textit{value}$ & who have a Stanford degree?\\
\hline
is-a noun    & $\times$                                        & $\times$                                          & alumni of $\textit{value}$, $\textit{value}$ alumni & who are alumni of Stanford?\\
\hline
active verb  & serves $\textit{value}$ cuisine                 & $\times$                                          & studied at $\textit{value}$, attended $\textit{value}$ & who attended Stanford?\\
\hline
passive verb & $\times$                                        & rated $\textit{value}$ star                       & educated at $\textit{value}$ & who are educated at Stanford?\\
\hline
adjective    & $\textit{value}$                                & $\textit{value}$-star                             & $\textit{value}$ & who are Stanford people?\\
\hline
prepositional& $\times$                                        & $\times$                                          & from $\textit{value}$ & who are from Stanford?\\
\bottomrule
\end{tabular}
\label{table:canonicals}
\end{table*}

\subsection{Generic Query Templates}

\paragraph{Sentence Types}
Let's start with the concept of sentence types. In English, there are four types: declarative, imperative, interrogative, and exclamatory. In \QandA, we care about declarative (``I am looking for $\ldots$''), imperative (``Search for $\ldots$'') and interrogative (``What is $\ldots$'').
We have generic query templates to generate all these three different sentence types. 
%Table~\ref{table:question_types2} shows examples for imperative and interrogative queries. 

Based on the type of the object in question, different interrogative pronouns can be used, in lieu of the generic ``what''.
% ``Who, what, where, when, why'' are 5 common pronouns in real life. Ignoring ``why'', which is not typically answered with a database, the rest are all mapped to the ``what'' questions in relational algebra. 
``Who'' maps to ``what is the person that''; ``where'' maps to ``what is the location of''; ``when'' maps to ``what is the time that''.   The distinction between persons, animals/inanimate objects, locations, and time is so important to humans that it is reflected in our natural language.  To create natural-sounding sentences, we create generic query templates for these different ``W''s and select the one to use according to the declared types of the fields in sentence synthesis. 

In addition, we can ask {\em yes-no questions} to find out if a stated fact is true, or 
{\em choice} questions, where the answer is to be selected from given candidates. 

%Move to the experimentation section.  Empirically, we found these two types of questions to be much less common in the domains we experiment on. 
%The templates for them are left for future work. 

\paragraph{Question Structure}
There is a great variety in how questions can be phrased, as illustrated by questions on the simple fact that ``Dr. Smith is Alice's doctor'' in Table~\ref{table:question_types2}.
We create templates to combine the following three factors to achieve variety in question structure. 
\begin{itemize}
\item
Two-way relationships.  Many important relationships have different words to designate the two parties; e.g. doctor and patient, employer and employee. 
\item 
  Parts of speech. A relationship can be expressed using phrases in different parts of speech.  Besides has-a noun, is-a noun, active verb, and passive verb, as shown in Table~\ref{table:question_types2}, we can also have adjective and prepositional phrases. Not all fields can be expressed using all the different parts of speech, as shown for ``servesCuisine'' and ``rating'' in Table~\ref{table:canonicals}. 
% preposition, and adjective (Table~\ref{table:canonicals}). 
%In addition to the canonical noun phrase used by templates in Table~\ref{table:db-ops}, we add two different noun phrases that describe what the subject has (has-a noun phrase) vs. what the subject is (is-a noun phrase).
\item
  Information of interest. We can ask about either party in a relationship; we can ask for Alice's doctor or Dr. Smith's patient.
\end{itemize}

In our knowledge representation, the rows in each table represent unique entities, each with a set of fields. 
Suppose we have a table of patients with ``doctors'' as a field; asking for Alice's doctor maps to a projection operation, whereas asking for Dr. Smith's patient maps to a selection operation.  
On the other hand, with a table of doctors with ``patients'' as a field, the converse is true.  
To avoid having multiple representations for the same query, our semantic parser assumes the existence of only one field for each relationship. 
It is up to the implementation to optimize execution if multiple fields exist.

{\renewcommand{\arraystretch}{0.9}
\begin{table}[tb]
\small
\caption{Type-specific comparison words.}
\vspace{-0.5em}
\label{table:type-comparisons}
\begin{tabular}{ll}
\toprule
\bf Type & \bf Comparative \\
\midrule
Time     & earlier, before, later, after \\
Duration & shorter, longer \\
Distance & closer, nearer, farther, more distant \\
Length & shorter, longer \\
Currency & cheaper, more expensive \\ 
Weight & lighter, smaller, heavier, larger \\
Speed & slower, faster \\
Temperature & colder, hotter \\
\bottomrule
\end{tabular}
\vspace{-0.5em}
\end{table}
}

\paragraph{Types and Measurements}
\iffalse
``Who, what, where, when, why'' are 5 common kinds of questions asked in real life. Ignoring ``why'', which is not typically answered with a database, the rest are all mapped to the ``what'' questions in relational algebra. 
``Who'' maps to ``what is the person that''; ``where'' maps to ``what is the location of''; ``when'' maps to ``what is the time that''.   The distinction between persons, animals/inanimate objects, locations, and time is so important to humans that it is reflected in our natural language.  To create natural-sounding sentences, we create generic query templates for these different ``W''s and select the one to use according to the declared types of the fields in sentence synthesis.  
\fi

The encoding of types in language carries over to measurements. Mathematical operators such as ``$a < b$'', ``$a > b$'', ``min'', ``max'' translate to different words depending on the kind of measurements. For example, for weight we can say ``heavier'' or ``lighter''. The list of type-specific comparison words is shown in Table~\ref{table:type-comparisons}. For each type, there are corresponding superlative words (``heaviest'', ``lightest'', etc.).
There is a relatively small number of commonly used measurements.  By building these varieties in the training data synthesis, we do not have to rely on manually annotated data to teach the neural networks these concepts. 

Types are also expressed in certain common, shortened question forms. For example, rather than saying ``what is the distance between here and X'', we can say ``how far is X''; the ``distance'' function and the reference point ``here'' are implicit. Similarly, instead of saying ``the restaurant with the shortest distance from here'', we can say ``the nearest restaurant''. These questions are sufficiently common that it is useful to include them in the generic set.

\iffalse
\begin{verbatim}
Length: shorter, longer, shortest, longest
Weight: lighter, heavier, lightest, heaviest
Time: earlier, later, earliest, latest
Cost: cheaper, more expensive, cheapest, most expensive
\end{verbatim}
\fi

\iffalse
\begin{table*}
\caption{Different part-of-speech for modifiers.}
\small
\begin{tabular}{l|l|l|l}
\toprule
{\bf Synthesis POS} & {\bf Example field} & {\bf Example phrase} & {\bf Example sentence}\\
\midrule
has-a noun phrase & alumniOf & degree from \# &  show me people who have a degree from X university\\
\hline
is-a noun phrase & alumniOf & alumni of \# & show me people who are alumni of X university\\
\hline
verb phrase & alumniOf & graduated from \# & show me people who graduated from X university  \\
\hline
passive verb phrase & alumniOf & educated at \# & show me people who are educated at X university\\
\hline
adjective phrase & servesCuisine & \# & show me restaurants which is Italian.\\
\hline
preposition phrase & geo & in \# & show me restaurants which is in X city.\\
\bottomrule
\end{tabular}
\label{table:pos}
\end{table*}
\fi
\iffalse
\TODO{Make into a table, add examples, and description}
\begin{verbatim}
Implicit subject: <property>
Adjective:        <property><subject>
                  <argmax/argmin><subject>
Phrase: With <property> <operator> <value>
        <passive-verb><value>
Clauses with variety in types of subject: which have/which are, who are, whose 
             adjective, 
             passive voice, 
             implicit "fields",  
             argmax table
\end{verbatim}
\fi

\paragraph{Connectives}
In natural language, people use connectives to compose multiple clauses
together, when they wish to search with multiple filters or joins.
Clauses can be connected with an explicit connective such as ``and'', ``but'', and ``with''. 
However, the connective can also be implicit, when composing natural short phrases 
based on its part-of-speech category.
%The different POS categories and connectives give modifiers great varieties in natural language.
%By taking into account what modifiers are used by two phrases, we can combine them in
%a natural way. % Fig.~\ref{fig:connectives} shows some example of possible combinations.
For example, multiple adjective phrases can be concatenated directly: 
people say ``5-star Italian restaurants'' rather than ``5-star and Italian restaurants''.

\subsection{Supplying Domain-Specific Knowledge}
The domain-specific information necessary to generate questions is provided by the developer
in the form of annotations on each field. Table~\ref{table:canonicals} shows examples of annotations for 3 fields
in the database. The developer is asked to provide common variations for each property, and they 
can iterate the annotations based on error analysis.
Even for the same POS category, there may be alternatives using different words.
For example, ``alumniOf'' can have two different verb phrases ``studied at'' and ``attended''.
The examples in this table also show that not every field can be referred to using the 6 POS modifiers.  
For restaurants, 
we can say ``has French food'', ``serves French food'', ``French restaurant'', but not
``restaurant is a French'', ``served at French'', or ``a restaurant of French''. 
Similarly, for ratings, we can say ``has 5 stars'', ``rated 5 stars'', ``5-star restaurant'', but not ``restaurant is a 5 rating'', ``restaurant rates 5'', or a ``restaurant of 5 stars''.  
We rely on the domain expert to annotate their fields with appropriate phrases for each POS category.  

\subsection{Template-Based Synthesis}

To synthesize data for the neural semantic parser, \webqa uses the Genie template-based 
algorithm~\cite{geniepldi19}. 
Genie expands the templates by substituting the non-terminals with the previously generated
derivations and applying the semantic function. Because the expansion is exponential, only a limited
fraction of the possible combinations are sampled. The expansion continues recursively, up to a configurable depth. 

A small portion of the synthesized data is paraphrased by crowdsource workers.
Both the synthetic and paraphrased sets are then augmented by replacing the field values with 
the actual values in the knowledge base. 
This allows us to expose more real values to the neural network
to learn how they match to different fields.

Because synthesized data is used in training, \webqa training sets cover a lot more questions than previous methods based
only on paraphrasing. \webqa does not need to rely on the distribution of sentences in the training set to match the test set.
Instead, it only needs to generalize minimally from sentences that are present
in the training set. As long as the training set has good coverage of real-world questions, this ensures high accuracy.
Additionally, developers can refine their skills with more templates.

\iffalse
\TODO{move down} The pipeline to get the final training examples is shown in Figure~\ref{fig:genie}.

\begin{figure}[t]
\centering
\includegraphics[width=0.9\linewidth]{./figures/genie-diagram.pdf}
\caption[]{Training data synthesis pipeline.}
\label{fig:genie}
\Description{}
\end{figure}
\fi

%!TEX root = ../paper.tex
\section{Q\&A for Schema.org}
\label{sec:representation}

\webqa is applicable to any \textit{ontology}, or database schema, that uses strict types with a type hierarchy consisting of the base classes of people, things, places, events.
To answer questions on the structured web, we develop NL-schema Builder to convert the RDF-based Schema.org ontology into an NL-schema.

\iffalse
into the annotated relational database schema used by the Thingpedia skill library, using the Thingpedia Schema Builder tool we developed. 

One important such ontology is Schema.org, a commonly used specification for representing structured information in web pages. 

By applying \webqa to Schema.org, we can immediately answer
questions on existing web pages. In this section, we first introduce Schema.org, then discuss
how the Schema.org ontology is adapted to \webqa, and the additional information that must
be provided in each domain to build high-quality Q\&A for that domain.
\fi

\subsection{Data Model of Schema.org}
% do not change this back, nodes are not classes, nodes *belong to* classes
Schema.org is a markup vocabulary created to help
search engines understand and index structured data across the web. 
It is based on RDF, and uses a graph data model, where nodes represent objects. Nodes are connected by \textit{properties}
and are grouped in \textit{classes}.
Classes are arranged in a multiple inheritance hierarchy where each class can be a subclass of 
multiple other classes, and the ``Thing'' class is the superclass of all classes.
% There is also a parallel hierarchy where the literal data types are defined.
% By convention, all class names start with an uppercase letter, while property names start with a lowercase letter.
%domain is just the regular maths domain, not a defined term

Each property's domain consists of one or more classes that can possess that property.
 The same property can be used in separate domains. For example, ``Person'' and ``Organization'' classes both use the ``owns'' property.
 % Subclasses of a class in the domain of a certain property can also make use of that property.
Each property's range consists of one or more classes or primitive types.  
Additionally, as having any data is considered better than none, the ``Text'' type (free text) is always implicitly included in a property's range.  For properties where free text is the recommended or only type expected, e.g. the ``name'' property, ``Text'' is explicitly declared as the property type. 

\subsection{NL-schema Builder}
To adapt the Schema.org ontology for use by \webqa, the NL-schema Builder (1)
converts the graph representation into database tables, (2) defines their fields and assigns the types, and (3)
provides annotations for each field. 

\iffalse
\webqa includes a converter tool
that leverages the Schema.org definitions and the data in the target website to perform (1) and (2)
automatically; the developer then provides (3). An example of a converted domain in Schema.org is shown in Fig.~\ref{fig:example-schema}

\begin{figure}[tb]
\begin{lstlisting}
skill Restaurant {
  Thing(
    name: String #_[prop=["name"]],
    image: Picture #_[prop=["image", "picture"]],
    description: String #_[prop=["description"]], ...
  );
  Place extends Thing(
    aggregateRating: {                      
      reviewCount: Number #_[prop=["review count"]],
      ratingValue: Number #_[
        prop=["rating"],
        passive_verb=["rated # star"}] 
      ]
    }, ... 
  );
  Organization extends Thing();
  LocalBusiness extends Place, Organization();
  FoodEstablishment extends LocalBusiness(
    servesCuisine: String #_[ 
      prop=["cuisine", "food type"], 
      verb=["serves # cuisine", "offer # food"] 
    ], ...
  );
  Restaurant extends FoodEstablishment();
}

\end{lstlisting}
\caption{The Restaurant domain, converted to the \webqa representation.}
\label{fig:example-schema}
\end{figure}
\fi

\paragraph{Creating tables.}
We categorize classes in Schema.org as either {\em entity} or {\em non-entity} classes. 
Entity classes are those that refer to well-known identities, such as people, organizations, places, events, with names that the user can recognize.
The Builder identifies entity classes using the hierarchy in Schema.org, and converts each class into a table in the \webqa representation. The properties of entity classes are converted into fields of the table. Given a target website for the \QandA system, the Builder 
only includes those properties used by the website to eliminate irrelevant terms.

Non-entity classes can only be referred to as properties of entity classes.  The Builder inserts
properties of non-entity classes directly into the entity class that refers to them, giving them unique names with a prefix. 
This design eliminates tables that cannot be referred to directly in natural language, hence simplifying synthesis. It also reduces the number of joins required to query the database, hence simplifying translation.

\iffalse
%Note the converse is not true: a property can refer to an entity class. 
This distinguish between entity and non-entity classes matches
how each class is used in natural language, and simplifies the query representation. 

An entity class is represented as a table, whose properties make up the columns.  
The column may be of a primitive type, a reference to another entity class, or a record type.
Non-entity classes are represented as anonymous record types.
In practice, most classes only use a subset of their properties.
\fi

\iffalse
\webqa does not allow recursive record types, where a record type has a field with the same type.  
Recursive non-entity classes are mapped to nameless tables, instead of record types. 
%\TODO{Can't follow: simply you are saying a review can be reviewed?
% TODO ANSWER: yes
For example, the ``Review'' class is a non-entity class, because it inherits the ``review'' property from ``CreativeWork'', and that property also refers to the ``Review'' class.
\fi

\paragraph{Assigning field types.}
The Schema.org class hierarchy provides most of the information needed by \webqa. 
However, the union type in Schema.org is problematic: it is not supported in \webqa to reduce ambiguity when parsing natural language. Here, the Builder trades off precision for ease of translation.  For each property, the Builder simply picks among the types in its range the one with the highest priority. The types in decreasing order of priority are: record types, primitive types, entity references, and finally strings. All the website data are cast to the chosen type.

\webqa needs to know the cardinality of each property, but that is not provided by Schema.org.
The Builder uses a number of heuristics to infer cardinality from the property name, the type, as well as the documentation. Empirically, we found the heuristics work well in the domains we evaluated, with the exception of properties of the ``Thing'' class, such as ``image'' and ``description'', which are described as plural in the documentation, but have only one value per object in practice.

\paragraph{Generating per-property natural language annotations.}
The Builder automatically generates one annotation for each property, based on the property name
and type. Camel-cased names are converted into multiple words and redundant words at the beginning and end are removed. The Builder uses a POS tagger and heuristics to identify the POS of the annotation. While the Builder will create a basic annotation for each property, developers are expected to add other annotations to improve the quality of synthesized sentences. 

%\input{sections/representation-compressed}
%!TEX root = ../paper.tex
\section{Neural Semantic Parsing Model}
\label{sec:model}

\begin{figure}
\includegraphics[width=0.9\linewidth]{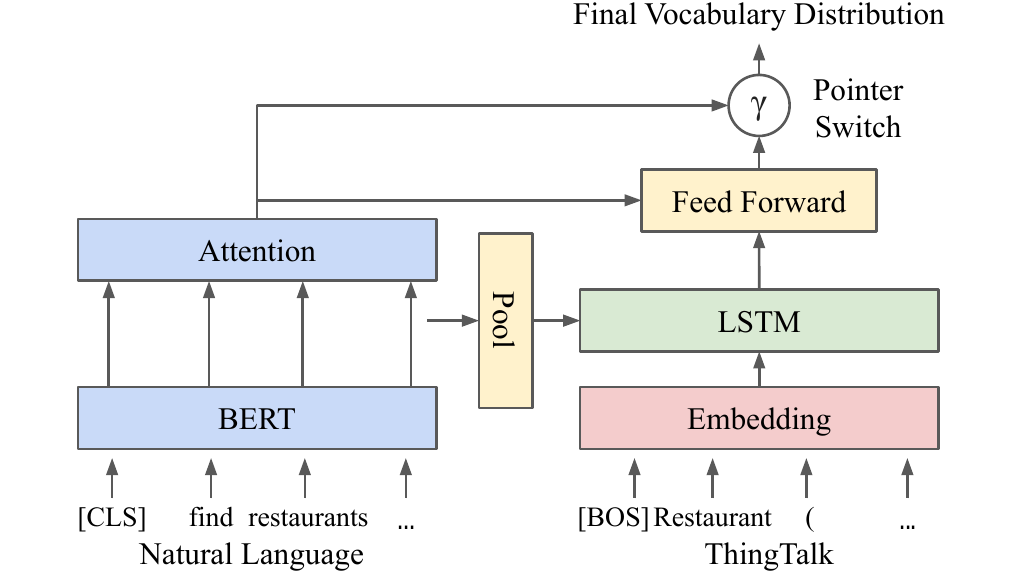}
\vspace{-0.5em}
\caption{The BERT-LSTM model in \SQA.}
\vspace{-0.8em}
\label{fig:model}
\end{figure}

\SQA uses a neural semantic parser to translate the user's question
to an executable query in \DBTalk. Our model is a simple yet novel architecture
we call BERT-LSTM. It is an encoder-decoder architecture
that uses the BERT pretrained encoder~\cite{Devlin2018Oct} and an LSTM
decoder with attention and a pointer-generator~\cite{bahdanau2014neural, jia2016data, see2017get}. This section introduces
the model and the rationale for its design. The overall architecture of the model
is shown Fig.~\ref{fig:model}.

\subsection{Encoding}
Our model utilizes the BERT model~\cite{Devlin2018Oct} as the sentence encoder.
We designed our encoder with the minimal amount of additional parameters on top of BERT, so as
to make the most use of pretraining.

BERT is a deep Transformer network~\cite{vaswani2017attention}. It encodes the sentence by first splitting
it into \textit{word-piece} subtokens,
then feeding them to a 12-layer Transformer network, to compute the final
contextualized dense representations of each token $h_{\text{E}}$.
BERT is pre-trained on general English text using
the \textit{masked language model} objective. We fine-tune it on our semantic parsing task.

To compute a single vector representation $\bar{h}_{\text{E}}$ for the whole sentence, the token representations produced by BERT are averaged, then fed to a one-layer feed-forward network:
\begin{align*}
h_{\text{E}} &= \text{BERT}(x)\\
\bar{h}_{\text{E}} &= W_{\text{pool,out}} \text{relu}(W_{\text{pool,in}} \text{avg}(h_{\text{E}}) + b_{\text{pool}}),
\end{align*}
where $x$ is the input sentence and $W$ and $b$ are learnable parameters.

% To improve learning and regularization, normalization and dropout layers are added around the pooling layer.

\subsection{Decoding}

During decoding, the model produces one token of the executable query $y_t$ at time $t$, given the previously
produced token $y_{t-1}$. There are no syntax-specific decoder components, and the model
can generalize to any target language, including future extensions of \DBTalk, depending on the training data.

First, the previous token is embedded using a learned embedding.
The embedded token is fed to an LSTM cell to compute the decoder representations $h_{\text{D},t}$. These
are then used to compute the attention scores $s_t$ against each token in the encoder, and the attention value
vector $v_t$ and the attention context vector $c_t$:
\begin{align*}
h_{\text{D},0} &= \bar{h}_{\text{E}}\\
c_{0} &= \mathbf{0}\\
y_{\text{emb},t} &= W_{\text{emb}} y_{t-1}\\
h_{\text{D},t} &= \text{LSTM}(h_{\text{D},t-1}, \left[y_{\text{emb},t} ; c_{t-1}\right])\\
%\end{align*}
%\begin{align*}
s_{t} &= \text{softmax}(h_{\text{D},t} h_{\text{E}}^{\text{T}})\\
v_{t} &= \sum_{t'} s_{t,t'} h_{\text{E},t'}\\
c_{t} &= \tanh(W_{\text{att}} \left[ v_t ; h_{\text{D},t} \right])
\end{align*}
(``emb'' denotes embedding and ``att'' denotes attention).
%(``$;$'' indicates concatenation).

The model then produces a vocabulary distribution $p_{t,w}$, where $w$ is the word index in the vocabulary. $p_{t,w}$ is either a distribution over tokens in
the input sentence (using a \textit{pointer} network~\cite{vinyals2015pointer}), or a distribution over the vocabulary.
The use of a pointer network allows the model to be mostly agnostic to specific entity names mentioned
in the question, which are copied verbatim in the generated query. This allows the model to generalize
to entities not seen in training. Conversely, the generator network allows the model to understand
paraphrases and synonyms of the properties, as well as properties that are only mentioned implicitly.

We employ the pointer-generator network previously proposed by \citeauthor{see2017get}~\cite{see2017get}.
The choice of whether to point or to generate from the vocabulary is governed by a
switch probability $\gamma_t$:
\begin{align*}
\gamma_t &= \sigma(W_{\gamma} \left[y_{\text{emb},t} ; h_{\text{D},t} ; c_{t}\right])\\
p_{t,w} &= \gamma_t \sum_{{t'}, x_{t'} = w} s_{t,t'} + (1 - \gamma_t) \text{softmax} (W_{\text{o}} c_{t})\\
y_t &= \arg\max_w p_{t,w}
\end{align*}
where $W_{\text{o}}$ is the output embedding matrix.

The model is trained to maximize the likelihood of the query for a given question, using
teacher forcing. At inference time, the model greedily chooses the token with the highest probability.

The model predicts one token of the query at a time, according to the tokenization rules of \DBTalk.
To be able to copy input tokens from the pretrained BERT vocabulary, all words between quoted strings
in the \DBTalk query are further split into word-pieces, and the model is trained to produce individual
word-pieces.

\iffalse
\subsection{Discussion}
State-of-the-art models for semantic parsing often use syntax-driven decoding, in which the model
is trained to predict AST nodes rather than left-to-right tokens~\cite{yin2017syntactic, rabinovich2017abstract, guo2019towards}.
These designs are inflexible, and require a syntax-specific intermediate representations. In \SQA, the model is trained with very large
corpus of automatically generated queries, and BERT-LSTM model can
learn the simple syntax of code quite easily.
\fi

\iffalse
%The choice of decoder affects the convergence speed during training. 
The LSTM network in \SQA can learn the syntax in a few iterations. 
We experimented with a 
Transformer decoder~\cite{vaswani2017attention} and found it slow in learning the 
correct syntax of \DBTalk.  It is important for the training to converge quickly, as too many updates to the encoder would reduce the
effectiveness of pretraining.
\fi

%!TEX root = ../paper.tex
\section{Experimental Results}
\label{sec:experiments}
\iffalse
We used \SQA to create three QA skills in the open-source Almond virtual assistant~\cite{almondwww17}, as an end-to-end demonstration of its functionality.
We used three aggregator websites (Yelp, LinkedIn and Hyatt Hotels) to build the skills. We then extended the skills with 311 hotel websites and 475 restaurant
websites that we crawled.
% then applied to 
% To validate this capability, we created two Q\&A systems, one for restaurants and one for hotels, in the cities of Washington DC and New York, and in the state of Hawaii.  
% We use Google Custom Search Engine and identify 311 hotel and 475 restaurant websites that include Schema.org markup.
\fi
%At the moment, our system supports only American English questions, 
%we expect the same methodology should work for different languages 
%with a different set of templates and natural language annotations.
%The skills and \SQA will be released upon publication. 

% We implemented an end-to-end prototype of \SQA, using the Genie toolkit~\cite{geniepldi19} for data synthesis and the decaNLP library~\cite{McCann2018decaNLP} for the neural model. %; \SQA generates skills for the Almond virtual assistant~\cite{almondwww17}.

In this section, we evaluate the performance of \SQA with experiments to answer the following:
%We first describe the dataset we used for training and evaluation, then describe the experiments to answer the following:
(1) How well does \SQA perform on popular Schema.org domains?
(2) How does our neural network model compare with prior work? 
(3) How do our templates compare with prior work? 
(4) Can the knowledge learned from one domain be transferred to a related domain?
(5) How do skills built with \SQA compare with commercial assistants?

As \webqa generates queries to be applied to a knowledge base, we evaluate on \textit{query accuracy}, which measures if the generated query matches the \DBTalk code exactly. 
The answer retrieved is guaranteed to be correct, provided the database contains the right data. 

In all the experiments, we use the same generic template library, which was refined over time, totally approximately 800 templates. It covers all the varieties described in Section~\ref{sec:nl-to-dbtalk}, except choice and yes-no questions.  Our experiments are conducted in English. Generalization to other languages is out of scope for this paper.

%To evaluate the performance on queries of different complexity, we further divide the dev and test according to the
%number of properties used in \webtalk query (including computed fields such as ``distance'' and ``count'').

We experiment on five Schema.org domains: restaurant, people, movies, books, and music.
We scrape Schema.org metadata from Yelp, LinkedIn, IMDb, Goodreads, and Last.fm, for each domain, respectively.  We include (1) all properties of these domains in Schema.org for which data is available, and (2) useful properties of types which no data is needed for training, such as enum and boolean type. 
The number of properties used per domain is shown in Table~\ref{table:evalset-size}.
We end up writing about 100 annotations per domain, many of which result from error analysis of the validation data. 

\begin{table}
\small
\caption{Size of training, dev, and test sets.}
\vspace{-0.5em}
\begin{tabular}{llrrrrr}
\toprule
&  \multicolumn{2}{r}{\bf Restaurants} & {\bf People} & {\bf Movies} & {\bf Books} & {\bf Music}\\
\midrule
\multicolumn{2}{l}{\# of properties}  & 25 & 13 & 16 & 15 & 19 \\
\multicolumn{2}{l}{\# of annotations} & 122 & 95 & 111 & 96 & 103 \\
\midrule
\multirow{3}{*}{Train} & Synthesized     & 270,081  &  270,081 & 270,081 & 270,081 & 270,081    \\
                       & Paraphrase      &  6,419   &  7,108   & 3,774   & 3,941   & 3,626\\
                       & {\bf Augmented} &  508,101 &  614,841 & 405,241 & 410,141 & 425,041\\
\midrule
\multirow{4}{*}{Dev} & 1  property    & 221      & 127     & 140     & 107     & 62      \\
                     & 2  properties  & 219      & 346     & 226     & 222     & 182     \\
                     & 3+ properties  & 88       & 26      & 23      & 33      & 82      \\

                     & \bf Total      & \bf 528  & \bf 499 & \bf 389 & \bf 362 & \bf 326 \\
\midrule
\multirow{4}{*}{Test} & 1  property   & 200      & 232     & 130     & 114     & 44      \\
                      & 2  properties & 245      & 257     & 264     & 241     & 181     \\
                      & 3+ properties & 79       & 11      & 19      & 55      & 63      \\

                      & \bf Total     & \bf 524  & \bf 500 & \bf 413 & \bf 410 & \bf 288 \\
\bottomrule
\end{tabular}
\label{table:evalset-size}
\end{table}
\begin{figure}
\centering
\includegraphics[width=0.95\linewidth,trim={0 0 0 0.45cm},clip]{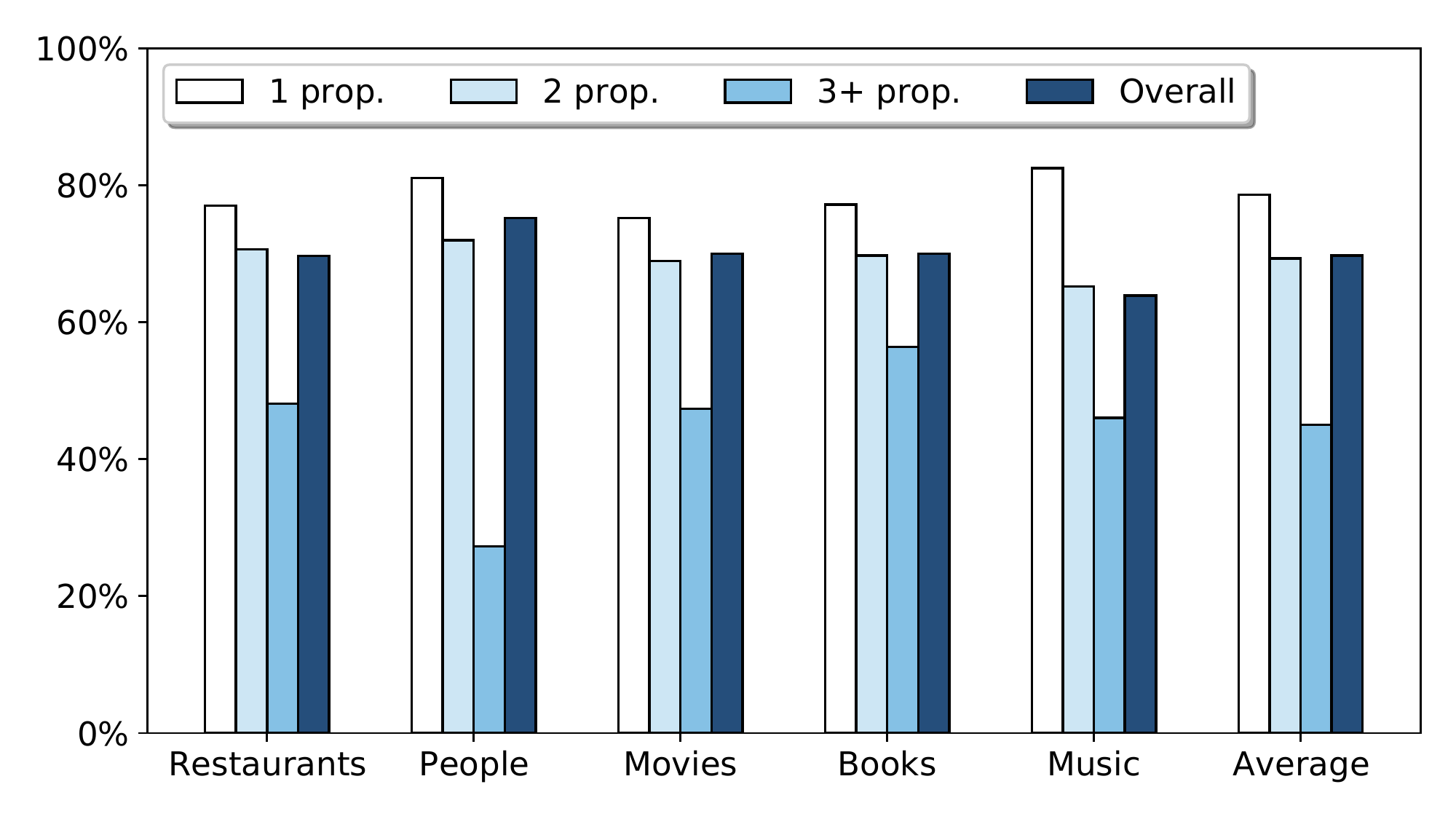}
\vspace{-1em}
\caption[]{Query accuracy on crowdsourced questions on Schema.org domains.}
\label{fig:accuracy-main}
\Description{}
\end{figure}
\subsection{\SQA on Five Schema.org Domains}
\label{sec:expt2domains}

\iffalse
Restaurants and people are two popular domains that use Schema.org meta-data extensively. 
The popular Yelp restaurant aggregator uses 10 of the Schema.org properties including 
``servesCuisine'', ``reviews'', ``aggregateRating'', 
as well as reviews with 4 properties: ``reviewRating'', ``author'', ``dataPublished'', and ``description''.  The popular LinkedIn site uses 5 of the Schema.org properties on people: ``alumniOf'', ``worksFor'', ``address'', ``award'', and ``name''.  Our first experiment evaluates the accuracy of skills we created by scraping these two websites and applying \SQA.   A screenshot
of the interface for the Yelp skill in the open-source Almond virtual assistant~\cite{almondwww17} is shown in Fig.~\ref{fig:screenshot}.  
\SQA composes the answer based on the return values of the Yelp skill 
and displays images when available.

\begin{figure}
\includegraphics[width=0.45\linewidth]{figures/sf_almond.png}
\vspace{-0.8em}
\caption{Screenshot of a restaurant skill generated by \SQA, running in the Almond virtual assistant.}
\vspace{-1em}
\label{fig:screenshot}
\end{figure}
\fi

The size of the generated training data and the evaluation data is shown in Table~\ref{table:evalset-size}. No real data are used in training. Furthermore, only less than 2\% of the synthesized data are paraphrased by crowdsource workers, resulting in a low data acquisition cost. Realistic data, however, are used for validation and testing, as they have been shown to be significantly more challenging and meaningful than testing with paraphrases~\cite{geniepldi19}.  

For the dev and test data, crowdsource workers are presented with a list of properties in the target domain and 
a few examples of queries, 
and are asked to come up with 5 questions with either one property or two properties, with equal probability.  
A property is counted once for every mention in the query. E.g., ``rating between 2 and 4''
would be counted as two properties, because it is equivalent to ``rating less than 4 and rating greater than 2''.
We observe that crowdsource workers generate questions that refer to three or more properties, despite being instructed to generate queries involving one or two. 
We do not show the workers any sentences or website data we used for training, and we allow them
to freely choose the value of each property. 
%Half of the workers are asked to produce queries about a single property, and the other half are asked queries related to two properties of their choice. %  Despite the instructions, some workers generate questions that use more than two properties.   
The questions are then annotated by hand with their \webtalk representation. The author who annotated the test set did not help tune the system afterwards. 

\iffalse
\begin{table}
\small
\caption{Training set sizes.}
\begin{tabular}{lrr}
\toprule
& {\bf Restaurants} & {\bf People}\\
\midrule
Synthesized  & 1,294,278 & 553,067 \\
Paraphrase   &     6,288 &   6,000 \\
\midrule
\bf Total (augmented) & \bf 1,789,105 & \bf 1,045,102 \\
\bottomrule
\end{tabular}
\label{table:training-size}
\end{table}
\fi

\iffalse
Based on the development set, we refined the generic templates. New templates we found
include: projection on two properties (``what is the address and the telephone of ...?''), filters that use ``both''
(``who works for both Google and Amazon?'', ``what restaurant serves both ramen and sushi?''), comparisons that use ``or more''
(``restaurants with 4 stars or more''). We also used the development set to refine the annotations of the properties.
\fi
%Applying \SQA to Aggregator Sites}

We train and evaluate on each domain separately.
The accuracy is shown in Fig.~\ref{fig:accuracy-main}.
On average, \SQA has a 69.7\% overall accuracy, achieving 78.6\%, 69.3\%, and 45.0\% accuracy on questions with one, two, three or more properties, respectively. 
Overall, this result, achieved without any real data in training, shows that
\SQA can build an effective parser at a low cost. Additionally, developers can add more annotations to further increase the accuracy with little additional cost.
\SQA is able to achieve reasonably high accuracy for complex queries because of the large synthesized training set, including many combinations of properties.

\iffalse
On Yelp data, \SQA achieves 74\% overall accuracy. Breaking down by the number of properties, \SQA achieves 76\% on questions with one property,
75\% on questions with two properties, and 63\% on three or more.
On LinkedIn data, \SQA achieves 78\% overall: 83\% on questions with one property 
and 74\% on questions with two properties. 
\fi
%In the figure, the column for three or more properties is marked ``N/A'' because there is no test data.
 % and the model is robust due to the large amounts of synthesized data.

\begin{table}
\small
\caption{Comparison of query accuracy across models.}
\begin{tabular}{lrrrrrr}
\toprule
          & \bf Restaurants & {\bf People} & {\bf Movies} & {\bf Books} & {\bf Music}  & {\bf Avg}\\
\midrule
MQAN      &     65.1\%  & 53.0\%       & 62.0\%     & 59.3\%     & 51.0\%     & 58.1\%\\
BERT-LSTM & \bf 69.7\%  & \bf 75.2\%   & \bf 70.0\% & \bf 70.0\% & \bf 63.9\% & \bf 69.7\%\\
\bottomrule
\end{tabular}
\label{table:accuracy-model-comparison}
\end{table}

\begin{table}
\small
\caption{Query accuracy of BERT-LSTM trained with only data synthesized from SEMPRE and Schema2QA templates.}
\begin{tabular}{lrrrrrr}
\toprule
%\multicolumn{2}{r}{\bf Restaurants} & {\bf People} & {\bf Movies} & {\bf Books} & {\bf Music}  & {\bf Avg}\
&{\bf Restaurants} & {\bf People} & {\bf Movies} & {\bf Books} & {\bf Music}  & {\bf Avg}\\
\midrule
SEMPRE     &  3.5\%     & 1.6\%      & 4.8\%      &  2.2\%     & 0.7\%      & 2.6\%\\
Schema2QA  & \bf 63.2\% & \bf 66.8\% & \bf 63.2\% & \bf 49.8\% & \bf 57.3\% & \bf 60.1\%\\
\bottomrule
\end{tabular}
\label{table:accuracy-template-comparison}
\end{table}

\iffalse
\begin{table}
\small
\caption{Query accuracy by number of properties.}
\begin{tabular}{lrr}
\toprule
& {\bf Restaurants} & {\bf People}\\
\midrule
1  property   & $78.6\% \pm 2.5\%$      & $85.7\% \pm 1.6\%$ \\
2  properties & $75.4\% \pm 0.8\%$      & $74.2\% \pm 2.0\%$ \\
3+ properties & $64.5\% \pm 1.6\%$      & N/A  \\
\bf Total     & $\mathbf{75.0\% \pm 1.5\%}$  & $\mathbf{79.8 \% \pm 1.9\%}$ \\
\bottomrule
\end{tabular}
\label{table:accuracy-main}
\end{table}
\fi

\iffalse
\begin{table}
\small
\caption{Answer accuracy of commercial virtual assistants.}
\begin{tabular}{lrrr}
\toprule
& {\bf Alexa} & {\bf Google} & {\bf Siri}\\
\midrule
Restaurant test  & 37\% & 46\% & 51\%\\
\bottomrule
\end{tabular}
\label{table:commercial-va-comparison}
\end{table}
\fi

\subsection{Neural Model Comparison}

The previous state-of-the-art neural model capable of training on synthesized data and achieving good accuracy was MQAN (Multi-Task Question Answering Network)~\cite{McCann2018decaNLP, geniepldi19}. It is an encoder-decoder
network that uses a stack of self-attention and LSTM layers, and uses GloVe and character word embeddings.
In this section, we evaluate how BERT-LSTM performs in comparison to MQAN.

As shown in Table.~\ref{table:accuracy-model-comparison},
the use of BERT-LSTM improves query accuracy by 11.6\% on average.  
This shows that the use of pretraining in BERT is helpful to generalize
to test data unseen in training.

\subsection{Evaluation of \SQA Templates}
\iffalse
\begin{figure}
\centering
\includegraphics[width=0.9\linewidth]{./figures/accuracy-comparison.pdf}
\caption[]{Impact of templates}
\label{fig:accuracy-comparison}
\Description{}
\end{figure}
\fi

One of the contributions of \SQA is a more comprehensive generic query template set than what was proposed originally by Sempre~\cite{overnight}. Here we quantify the contribution by comparing the two. 
%\SQA though uses a more comprehensive template set for QA. Here we evaluate the effectiveness of our templates.
%significantly more sophisticated template language and synthesis generation algorithm.
%Here we evaluate the quality of our synthesized data set compared to the one generated using the Sempre language. 
We reimplement the Sempre templates using Genie, and apply the same data augmentation to both sets.
We train with only synthesized data, and evaluate on realistic data. 
The result is shown in Table~\ref{table:accuracy-template-comparison}.
On average, training with Sempre templates achieves only 2.6\% accuracy. 
On the other hand, training with only synthesized data produced with \SQA templates achieves 59.5\% accuracy.
This result shows that the synthesized data we generate matches our realistic test questions more closely.
Our templates are more tuned to understand the variety of filters that commonly appear in the test.
%On the other hand, Sempre templates are tuned for questions with many joins, which are not common in the domains we tested.
Furthermore, due to the pretraining, the BERT-LSTM model can make effective use of synthesized data
and generalize beyond the templates.
These two effects combined means we do not need to rely as much on expensive paraphrasing.

\begin{table*}
\small
\vspace{-0.5em}
\caption{Comparing \SQA with three major commercial assistants on 10 queries about restaurants, people, and hotels.}
\vspace{-0.5em}
\label{table:comparison}
\begin{tabular}{p{9cm}cccc}
\toprule
{\bf Query} & {\bf Google} & {\bf Alexa} & {\bf Siri} & {\bf \webqa}\\
\hline
Show restaurants near Stanford rated higher than 4.5 & $\times$ & $\times$ & $\times$ & {\bf \cmark}\\
\hline
Show me restaurants rated at least 4 stars with at least 100 reviews & $\times$ & $\times$ & $\times$ & {\bf \cmark}\\

\hline
What is the highest rated Chinese restaurant in Hawaii? & $\times$ & {\bf \cmark} & {\bf \cmark} & {\bf \cmark}\\
%\hline
%How many reviews mentioned good wine for iTalico? & $\times$ & $\times$ & $\times$ & $\times$ \\
\hline
How far is the closest 4 star and above restaurant? & $\times$ & $\times$ & $\times$ & {\bf \cmark}\\
\hline 
Find a W3C employee that went to Oxford & $\times$ & $\times$ & $\times$ & {\bf \cmark}\\
\hline
Who worked for both Google and Amazon? & $\times$ & $\times$ & $\times$ & {\bf \cmark}\\
\hline
Who graduated from Stanford and won a Nobel prize? & {\bf \cmark} & $\times$ & $\times$ & {\bf \cmark}\\
\hline
Who worked for at least 3 companies? & $\times$ & $\times$ & $\times$ & {\bf \cmark}\\
\hline
Show me hotels with checkout time later than 12PM & $\times$ & $\times$ & $\times$ & {\bf \cmark}\\
%\hline
%List hotels reviewed by Johnny Jet? & $\times$ & $\times$ & $\times$ & {\bf \cmark}\\
\hline
Which hotel has a swimming pool in this area? & {\bf \cmark} & $\times$ & $\times$ & {\bf \cmark}\\
\bottomrule
\end{tabular}
\end{table*}

\subsection{Zero-Shot Transfer Learning to Hotels}

Many domains in Schema.org share common classes and properties.  
%Is it possible to transfer the learning from one domain to another and create a semantic parser for a new domain without any manual acquisition of training data?  i.e., without manual annotations or paraphrases?  
Here we experiment with transfer learning from the restaurant domain to the hotel domain. Restaurants and hotels
share many of the same fields such name, location, and rating. 
The Hotel class has additional properties ``petsAllowed'', ``checkinTime'', ``checkoutTime'', and ``amenityFeature''.
For training data synthesis, we have manual annotations for fields that are {\em common} with restaurants, and for the rest, we use annotations automatically generated by \SQA.  The paraphrases for the restaurant domain are automatically transferred to the hotel domain 
by replacing words like ``restaurant'' and ``diner'' with ``hotel''.
We augment the training sets with data crawled from the Hyatt hotel chain. 

%\TODO{What about the validation set?}
We acquire an validation set of 443 questions and a test set of 528 questions, crowdsourced from MTurk, and annotated by hand. %These are divided in 443 for validation and 528 for test. % set of 181 questions
205 of the test questions use one property, 270 use two properties, and 53 use three or more. 
On the test set, the generated parser achieves an overall accuracy of 64\%. 
Note that about half of the test sentences use properties that are specific to the ``Hotel'' class.
This shows that it is possible to bootstrap a new domain, similar to an existing one, with no manual training data acquisition.
%This shows that the work from one domain can be transferred to a similar domain with no manual training data acquisition,
%achieving an accuracy sufficient to bootstrap the new domain.

SQA has also been used in a recent transfer-learning experiment\cite{campagna-etal-2020-zero} on the MultiWOZ multi-domain dialogue dataset~\cite{budzianowski2018large}. A model for each domain was trained by replacing in-domain real training data with synthesized data and data automatically adapted from other domains. The resulting models achieve an average of 70\% of the accuracy obtained with real training data.

\subsection{Comparison with Commercial Assistants}
For the final and most challenging experiment, we compare \SQA with commercial assistants on Restaurant domain. 
We can only evaluate commercial assistants on their {\em answer accuracy} by checking manually if the results match the question, since they do not show the generated queries.  Answer accuracy is an upper bound of query accuracy because the assistant may return the correct answer even if the query is incorrect.

In the first comparison, we use the test data collected in Section~\ref{sec:expt2domains}.  
As shown in Fig.~\ref{fig:accuracy-commercial}, \SQA gets the best result with a 
query accuracy of 69.7\%; Siri has a 51.3\% answer accuracy, Google Assistant is at 42.2\%, and 
Alexa is at 40.6\%.
Unlike \SQA, commercial assistants are tuned to answer the more frequently asked questions. 
Thus, we perform another comparison where we ask crowdsource workers to ask any questions they like about restaurants, without showing them the properties in our database. Note that we only test the systems on questions that can potentially be answered using the Schema.org data. We collected 300 questions each for the dev and test set. Removing questions that cannot be answered with our subset of Schema.org, 
we have a total of 137 questions for dev and 169 for test.
Out of the 169 test questions, 27 use one property, 89 use two, and 53 use three or more.
The accuracy of \SQA drops from 69.7\% in the first test to 59.8\% in the second test, because the unprompted questions use a wider vocabulary. 
Nonetheless, its performance matches that of Google assistant, about 7.1\% lower than Siri and 14.7\% higher than Alexa. 
%Alexa has a much poorer answer accuracy of 45.0\%. 

The answer accuracy of commercial assistants improves significantly in the second test.
This is because they are tuned to answer popular commands, and can answer the question correctly
even with limited understanding of the question. For example, by default they always return restaurants nearby.  
On the other hand, we measure query accuracy for \SQA, where the exact ThingTalk code needs to be predicted.
\SQA can also apply the same heuristics to the returned answer, and we expect our answer accuracy, if measured on a large knowledge base, would be higher.

\iffalse
\begin{figure}
\centering
\begin{subfigure}{\linewidth}
\centering
\includegraphics[width=0.8\linewidth]{./figures/accuracy-commercial-old.pdf}
\caption[]{Restaurant test collected with property list shown}
\label{fig:accuracy-commercial-old}
\Description{}
\end{subfigure}
\begin{subfigure}{\linewidth}
\centering
\includegraphics[width=0.8\linewidth]{./figures/accuracy-commercial-new.pdf}
\caption[]{Restaurant test collected with no prompt}
\label{fig:accuracy-commercial-new}
\Description{}
\end{subfigure}
\caption[]{Answer accuracy of commercial assistants compared with query accuracy of \webqa.}
\label{fig:accuracy-commercial}
\end{figure}
\fi

\begin{figure}
\centering
\includegraphics[width=0.95\linewidth]{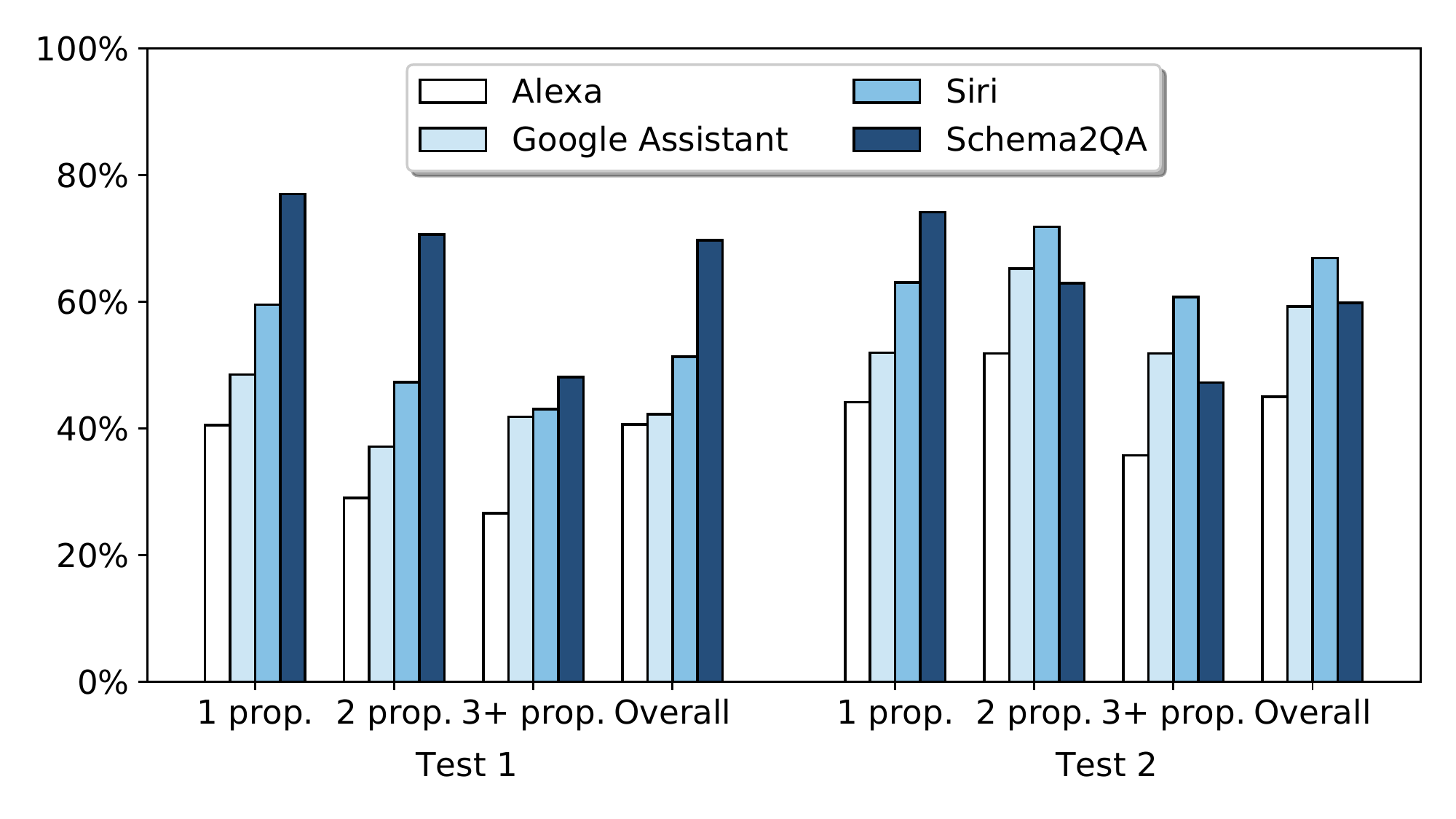}
\vspace{-1.2em}
\caption[]{Comparison of commercial assistants (answer accuracy) and \webqa (query accuracy) on Restaurant domain. In Test 1, question writers see available properties, and in Test 2 they do not.}
\vspace{-1.5em}
\label{fig:accuracy-commercial}
\end{figure}

\subsection{Discussion}

Error analysis on the validation set in our experiments of Section~\ref{sec:expt2domains} reveals that better named entity recognition can eliminate about half of the errors. An additional 14\% seems to be resolvable with additional generic templates, an example of which is selecting two fields with the same value (``movies produced and directed by ...''). Fixing these two issues can potentially bring the accuracy from 70\% to about 90\% for in-domain questions. The results confirm our hypothesis that natural-language queries can be factored into a generic and a domain-specific part. Our methodology enables developers to iteratively refine both the generic question templates and domain-specific annotations. Furthermore, by cumulating generic question knowledge in the open-source template library, we enable developers to bootstrap an effective agent for a new domain in just a few days without any real training data.

\section{Conclusion}
\label{sec:conclusion}
This paper presents \webqa, a toolkit for building question-answering agents for 
databases. \webqa creates semantic parsers that translate complex natural language questions involving multiple predicates and computations into \DBTalk queries.
By capturing the variety of natural language with a common set of generic question templates, we  
eliminate the need for manually annotated training data and minimize
the reliance on paraphrasing.
Developers only need to annotate the database fields with their types and a few phrases.
\webqa also includes an NL-schema Builder tool that adapts Schema.org to \SQA, and provides a complete pipeline to build Q\&A agents for websites.
%They can do so a domain at a time, with relatively little manual effort and cost.
%They need only to annotate the properties in Schema.org, and \webqa uses a comprehensive set of domain-independent templates to generate compound sentences and gets paraphrases for a small fraction of them.  

%\webqa uses \NLSchema, an extension of the Schema.org definitions that includes natural language canonical forms,
%to automatically generate a large data set to train the neural network. The training set is generated
%based on templates optimized for website Q\&A, in combination with canonical forms in 6 POS categories we identified.

Experimental results show that
our BERT-LSTM model outperforms the MQAN model by 4.6\% to 12.9\%. 
The Q\&A agents produced by \webqa can answer crowdsourced complex queries with 70\% average accuracy.  
On the restaurant domain, we show a significant improvement over Alexa, Google, and Siri, which can answer at most 51\% of our test questions.
Furthermore, on common restaurant questions that \webqa can answer, our model matches the accuracy of Google, and is within 7\% of Siri without any user data.
% With transfer learning, we show that a new domain can achieve a 65\% accuracy without manual effort.
% Furthermore, \webqa can answer 59\% of questions that use 3 or more properties.
By making \SQA available, we wish to encourage the creation of a voice web that is open to every virtual assistant.

\begin{acks}
We thank Ramanathan V.~Guha for his help and suggestions on Schema.org. 
This work is supported in part by the \grantsponsor{NSF}{National Science Foundation}{https://www.nsf.gov/awardsearch/showAward?AWD_ID=1900638&HistoricalAwards=false} 
under Grant No.~\grantnum{nsf}{1900638} and the \grantsponsor{Sloan}{Alfred P. Sloan Foundation}{} under Grant No.~\grantnum{sloan}{G-2020-13938}.
\end{acks}
 
\bibliographystyle{ACM-Reference-Format}
\bibliography{paper}

\end{document}